\documentclass[10pt,twocolumn,journal]{IEEEtran}

\usepackage{epsfig}
\usepackage{citesort}
\usepackage{amsmath}
\usepackage{amssymb}
\usepackage{color}
\usepackage{url}
\usepackage{multirow}
\usepackage{amsmath}
\usepackage{sub caption}
\usepackage{epstopdf}
\usepackage{cite}

\graphicspath{{Figures/}}

\newcommand{\argmin}{\operatornamewithlimits{argmin}}

\newlength{\figurewidth}
\newlength{\smallfigurewidth}

\begin{document}

\title
{
Composite Kernel Local Angular Discriminant Analysis for Multi-Sensor Geospatial Image Analysis
}

\author{%
Saurabh Prasad,~\IEEEmembership{Senior Member,~IEEE},
Minshan Cui,
Lifeng Yan
\thanks{%
The authors are with the Department of Electrical \&
Computer Engineering at the University of Houston. This material is based upon work funded in part by the NASA New Investigator (Career) Award, Grant No. NNX14AI47G. Corresponding author: Saurabh Prasad (email: saurabh.prasad@ieee.org).
}
}


\maketitle
\thispagestyle{empty}
\pagestyle{empty}


\begin{abstract}
With the emergence of passive and active optical sensors available for geospatial imaging, information fusion across sensors is becoming ever more important. An important aspect of single (or multiple) sensor geospatial image analysis is feature extraction --- the process of finding ``optimal'' lower dimensional subspaces that adequately characterize class-specific information for subsequent analysis tasks, such as classification, change and anomaly detection etc. In recent work, we proposed and developed an angle-based discriminant analysis approach that projected data onto subspaces with maximal ``angular'' separability in the input (raw) feature space and Reproducible Kernel Hilbert Space (RKHS). We also developed an angular locality preserving variant of this algorithm. In this letter, we advance this work and make it suitable for information fusion --- we propose and validate a composite kernel local angular discriminant analysis projection, that can operate on an ensemble of feature sources (e.g. from different sources), and project the data onto a unified space through composite kernels where the data are maximally separated in an angular sense. We validate this method with the multi-sensor University of Houston hyperspectral and LiDAR dataset, and demonstrate that the proposed method significantly outperforms other composite kernel approaches to sensor (information) fusion. 
\end{abstract}


\section{Introduction}
Optical remote sensing has made significant advances in recent years. Among these has been the deployment and wide-spread use of hyperspectral imagery on a variety of platforms (including manned and unmanned aircraft and satellites) for a wide variety of applications, ranging from environmental monitoring, ecological forecasting, disaster relief to applications pertaining to national security. With rapid advancements in sensor technology, and the resulting reduction of size, weight and power requirements of the imagers, it is also now common to deploy multiple sensors on the same platform for multi-sensor imaging. As a specific example, it is appealing for a variety of remote sensing applications to acquire hyperspectral imagery and Light Detection and Ranging (LiDAR) data simultaneously --- hyperspectral imagery offers a rich characterization of object specific properties, while LiDAR provides topographic information that complements Hyperspectral imagery \cite{dalponte2008fusion,YP2015,brennan2006object,shimoni2011detection,pedergnana2011fusion}. Modern LiDAR systems provide the ability to record entire waveforms for every return signal as opposed to providing just the point cloud. This enables a richer representation of surface topography. 

While feature reduction is an important preprocessing to analysis of single-sensor high dimensional passive optical imagery (particularly hyperspectral imagery), it becomes particularly important with multi-sensor data where each sensor contributes to high dimensional raw features. A variety of feature projection approaches have been used for feature reduction, including classical approaches such as Principal Component Analysis (PCA), Linear Discriminant Analysis (LDA) and their many variants, manifold learning approaches such as Supervised and Unsupervised Locality Preserving Projections \cite{lunga2014manifold}. Several of these methods are implemented in both the input (raw) feature space and the Reproducible Kernel Hilbert Space (RKHS) for data that are nonlinearly separable. 
Further, most traditional approaches to feature extraction are designed for single-sensor data --- a unique problem with multi-sensor data is that feature spaces corresponding to each sensor often have different statistical properties, and a single feature projection may hence be sub-optimal. It is hence desired to have a projection for feature reduction that preserves the underlying information from each sensor in a lower dimensional subspace. 

More recently, we developed a feature projection approach, referred to as Angular Discriminant Analysis (ADA) \cite{CP2015_ADA_JSTSP,PC2013Asilomar_Sparse,PrasadCuiICASSP2013}, that was optimized for hyperspectral imagery and demonstrated robustness to spectral variability. Specifically, the approach sought a lower dimensional subspace where classes were maximally separated in an angular sense, preserving important spectral shape related characteristics. We also developed a local variant of the algorithm (LADA) that preserved angular locality in the subspace. In this paper, we propose a composite kernel implementation of this framework and demonstrate for the purpose of feature projection in multi-sensor frameworks. Specifically, by utilizing a composite kernel (a dedicated kernel for each sensor), and ADA (or LADA) for each sensor, the resulting projection is highly suitable for classification. The proposed approach serves as a very effective feature reduction algorithm for sensor fusion --- it optimally fuses multi-sensor data and projects it to a lower dimensional subspace. A traditional classifier can be employed following this, for supervised learning. We validate the method with the University of Houston multi-sensor dataset comprising of Hyperspectral and LiDAR data and show that the proposed method significantly outperforms other approaches to feature fusion.

The outline of the remainder of this paper is as follows. In sec. \ref{sec:related}, we review related work. In sec. \ref{sec:proposed}, we describe the proposed approach for multi-sensor feature extraction. In sec. \ref{sec:results}, we describe the experimental setup and present results with the proposed method, comparing it to several state-of-the-art techniques to feature fusion.


\section{Related Work}
\label{sec:related}
Traditional approaches to feature projection based dimensionality reduction such as PCA, LDA and their variants largely rely on Euclidean measures. Manifold learning approaches \cite{lunga2014manifold} also seek to preserve manifold structures and neighborhood locality through projections that preserve such structures. Other projection based approaches to feature reduction, such as Locality Preserving Projections (LPP), Local Fisher's Discriminant Analysis (LFDA) \cite{lunga2014manifold,prasad2014SMoG,MPWB2013} etc. integrate ideas of local neighborhoods through affinity matrices, into classical projection based analysis approaches such as PCA, LDA etc. As a general feature extraction approach, Euclidean distance is a reasonable choice, including for remotely sensed image analysis. However, by noting the well understood benefits of spectral angle for hyperspectral image analysis, in previous work, we developed an alternate feature projection paradigm that worked with angular distance measures instead of euclidean distance measures \cite{CP2015_ADA_JSTSP,PC2013Asilomar_Sparse,PrasadCuiICASSP2013} --- we demonstrated that when projecting hyperspectral data through this class of transformations, the resulting subspaces were very effective for downstream classification and significantly outperformed their Euclidean distance counterparts. In addition to benefits with classification, we also demonstrated other benefits of this class of methods, including robustness to illumination differences --- something that is very important for remote sensing. In other previous work, it has been shown that a reproducible kernel Hilbert space (RKHS) generated by composite kernels (a weighted linear combination of basis kernels) is very effective for multi-source fusion \cite{YP2015,wang2013multiple,tuia2010multisource,camps2006composite}.
Here, we briefly review the developments related to angular discriminant analysis. This will provide a context and motivation for the proposed work in this paper that seeks to demonstrate the benefits of composite kernel angular discriminant analysis for multi-source image analysis. 

\subsection{Angular Discriminant Analysis}
Here, we briefly review Angular Discriminant Analysis (ADA) and its locality preserving counterpart, Local Angular Discriminant Analysis (LADA). Consider a $d$-dimensional feature space (e.g. hyperspectral imagery with $d$ spectral channels). Let $\{{x}_{i} \in \mathbb{R}^{d}, y_i \in \{1,2, \ldots, c\}\}$ be the $i$-th training sample with an associated class label $y_i$, where $c$ is the number of classes. The total number of training samples in the library is $n = \sum_{l=1}^{c}{n_{l}}$, where $n_{l}$ denotes the number of training samples from class $l$. Let ${\mathit{T}} \in \mathbb{R}^{d \times r}$ be the desired projection matrix, where $r$ denotes the reduced dimensionality. We also denote symbols having unit norm with a \emph{tilde} --- this will be useful where we normalize the data to a unit norm to focus on angular separability.

\subsubsection{ADA}
Traditional LDA seeks to find a subspace that maximizes between-class scatter while minimizing within-class scatter, where the scatter is measured using Euclidean distances. While similar in philosophy, ADA is an entirely new approach to subspace learning that is based on \emph{angular} scatter --- it seeks a subspace where within-class angular scatter is maximized, and the between-class angular scatter is maximized. Just like LDA, the ADA optimization problem can be posed as a generalized eigenvalue problem. Specifically, ADA seeks to find a projection where the ratio of between-class inner product to within-class inner product of data samples is minimized. The within-class outer product matrix ${\mathit{O}}^{(\text{w})}$ and between-class outer product matrix ${\mathit{O}}^{(\text{b})}$ are defined as
\begin{align}
{\mathit{O}}^{(\text{w})} &= \sum^{c}_{l=1}\sum^{}_{i:y_i=l} \tilde{{\mu}}_{l}\tilde{{x}}_{i}^{t},
\label{eq:Ow}
\\
{\mathit{O}}^{(\text{b})} &= \sum^{c}_{l=1}n_l\tilde{{\mu}}\tilde{{\mu}}_{l}^{t}.
\label{eq:Ob}
\end{align}  
{where $\tilde{{\mu}}_{l} = \frac{1}{n_l}\sum_{i:y_i=l}^{}{\tilde{{x}}_{i}}$ is the normalized mean of $l$-th class samples, and $\tilde{{\mu}} = \frac{1}{n}\sum_{i=1}^{n}{\tilde{{x}}_{i}}$ is defined as the normalized total mean.}

It was shown in \cite{CP2015_ADA_JSTSP} that the projection matrix ${\mathit{T}}_{\textit{ADA}}$ of ADA can be approximated as the solution to the following trace ratio problem
\begin{align}
{\mathit{T}}_{\textit{ADA}} 
&\approx \argmin_{{\mathit{T}}\in \mathbb{R}^{d \times r}}\left[\operatorname{tr}\big(({\mathit{T}}^{t}{\mathit{O}}^{(\text{w})}{\mathit{T}})^{-1}{\mathit{T}}^{t}{\mathit{O}}^{(\text{b})}{\mathit{T}}\big)\right].
\label{eq:ada2}
\end{align}
The projection matrix ${\mathit{T}}_{\textit{ADA}}$ can be obtained by solving the generalized eigenvalue problem involving ${\mathit{O}}^{(\text{w})}$ and ${\mathit{O}}^{(\text{b})}$. 


\subsubsection{LADA} 
Similar to LDA, ADA is a ``global'' projection in that it does not specifically promote preservation of local (neighborhood) angular relationships under the projection. We hence developed LADA in \cite{CP2015_ADA_JSTSP}, which is a local variant of ADA. The within and between-class outer product matrices of LADA are obtained as follows
\begin{align}
{\mathit{O}}^{(\text{lw})} &= \sum_{i,j=1}^{n}\mathit{\tilde{W}}_{ij}^{(\text{lw})}\tilde{{x}}_{i}\tilde{{x}}_{j}^{t}, \label{eq:lada_ow} \\
{\mathit{O}}^{(\text{lb})} &= \sum_{i,j=1}^{n}\mathit{\tilde{W}}_{ij}^{(\text{lb})}\tilde{{x}}_{i}\tilde{{x}}_{j}^{t},
\label{eq:lada_ob}
\end{align}
where the normalized weight matrices are defined as 
\begin{align}
\mathit{\tilde{W}}^{(\text{lw})}_{ij}&=
     \begin{cases}
          \mathit{\tilde{A}}_{ij}/n_l, & \text{if $y_i,y_j=l$}, \\
                    0, & \text{if $y_i\ne y_j$},
     \end{cases}
\label{eq:lada_w} \\
\mathit{\tilde{W}}^{(\text{lb})}_{ij}&=
     \begin{cases}
          \mathit{\tilde{A}}_{ij}(1/n-1/n_l), & \text{if $y_i,y_j=l$}, \\  
                         1/n, & \text{if $y_i\ne y_j$}.
     \end{cases}
     \label{eq:lada_b}
\end{align}

The normalized affinity matrix $\mathit{\tilde{A}}_{ij}\in [0,1]$ between $\tilde{{x}}_i$ and $\tilde{{x}}_j$ is defined as
\begin{equation}
\mathit{\tilde{A}}_{ij}=\exp\left(-\frac{( 2-2\tilde{{x}}_i^{t}\tilde{{x}}_j)}{\tilde{\gamma}_i\tilde{\gamma}_j}\right) ,
\label{eq:aff}
\end{equation}  
where $\tilde{\gamma}_{i}= \sqrt{2-2\tilde{{x}}_i^{t}\tilde{{x}}^{(\text{knn})}_i}$ denotes the \emph{local angular scaling} of data samples in the angular neighborhood of $\tilde{{x}}_i$, and $\tilde{{x}}^{(\text{knn})}_i$ is the
\emph{K}-th nearest neighbors of $\tilde{{x}}_i$. 

Similar to ADA, the projection matrix of LADA can be defined as
\begin{align}
{\mathit{T}}_{\textit{LADA}} &=  \argmin_{{\mathit{T}}\in \mathbb{R}^{d \times r}}\left[\operatorname{tr}\big(({\mathit{T}}^{t}{\mathit{O}}^{(\text{lw})}{\mathit{T}})^{-1}{\mathit{T}}^{t}{\mathit{O}}^{(\text{lb})}{\mathit{T}}\big)\right].
\label{eq:lada}
\end{align}



\section{Composite Kernel Angular Discriminant Analysis for Image Fusion}
\label{sec:proposed}
In this section, we develop and describe the proposed approach to multi-source feature extraction --- composite kernel angular discriminant analysis (CKADA) and its locality preserving counterpart (CKLADA). Our underlying hypothesis with this work is that even when angular information is important for optical image analysis, in a multi-source (e.g. multi-sensor scenario), having dedicated kernels (specific to each source) would result in a superior projection that addresses source-specific nonlinearities. With that goal, we extend our previous work with angular discriminant analysis by implementing it in a composite kernel reproducible kernel Hilbert space and demonstrate for a specific application of multi-sensor image analysis that the resulting subspace is highly discriminative and outperforms other subspace learning approaches. 

Consider a nonlinear mapping $\Phi(\cdot)$ from the input space to a RKHS $\mathcal{H}$ as follows: 

\begin{equation}
{\Phi}: \mathbb{R}^d \rightarrow \mathcal{H}, {x} \rightarrow {\Phi}({x}). 
\end{equation}

\noindent and a kernel function ${K}$ defined as:

\begin{equation}
{K}({{{x}}_i},{{{x}}_j}) = \left\langle {\Phi ({{{x}}_i}),\Phi ({{{x}}_j})} \right\rangle, 
\end{equation}
where $\left\langle {\cdot,\cdot} \right\rangle$ is the inner product of two vectors. Consider next a set of $M$ co-registered multi-source images resulting in the following $M$-Tuple of feature vectors from co-registered images for every geolocation (co-registered pixels): $\{{x}^1, {x}^2, \hdots, {x}^M\}$, where ${x}^m \in \mathbb{R}^{d_m}$. Associated with every pixel (geolocation) for which ground truth is available, there is a class label $y \in \{1, 2, \hdots, c\}$. A composite kernel RKHS can then be constructed as

\begin{equation}
{K}({x}_i, {x}_j) = \sum_{m=1}^M \alpha_m {K}_m ({x}^m_i, {x}^m_j)
\label{eq:kernelmix}
\end{equation}

\noindent where $K_m$ is a basis kernel for the $m$'th source, formed by any valid Mercer's kernel.
To implement Composite Kernel ADA (CKADA), note that ${\mathit{O}}^{(\text{w})}$ and ${\mathit{O}}^{(\text{b})}$ can be reformulated as
	\begin{align}
		{\mathit{O}}^{(\text{w})} &= {\mathit{X}}{\mathit{W}}^{(\text{w})}{\mathit{X}}^{t}, \\
		{\mathit{O}}^{(\text{b})} &= {\mathit{X}}{\mathit{W}}^{(\text{b})}{\mathit{X}}^{t}. 
	\end{align}
	
\noindent where ${\mathit{W}}^{(\text{w})}$ is given as
\begin{align}
\mathit{W}_{ij}^{(\text{w})}&=
     \begin{cases}
          1/n_{l}, & \text{if $y_i,y_j=l$},  \\
          0, & \text{if $y_i\ne y_j$}.
     \end{cases}
     \label{eq:ww2}
\end{align}

\noindent and ${\mathit{W}}^{(\text{b})}$ is given as
\begin{align}
\mathit{W}_{ij}^{(\text{b})}&=
     \begin{cases}
          1/n-1/n_l, & \text{if $y_i,y_j=l$}, \\
          1/n, & \text{if $y_i\ne y_j$}.
     \end{cases}
     \label{eq:wb2}
\end{align}
		
ADA can hence be re-expressed as the solution to the following generalized eigenvalue problem
\begin{align}
{\mathit{X}}{\mathit{W}}^{(\text{b})}{\mathit{X}}^{t}{\nu} = \lambda{\mathit{X}}{\mathit{W}}^{(\text{w})}{\mathit{X}}^{t}{\nu}.
\label{eq:keig}
\end{align}

Since ${\nu}$ can be represented as a linear combination of columns of ${\mathit{X}}$, it can be formulated using a vector ${\varphi} \in \mathbb{R}^{n}$ as
\begin{align}
{\mathit{X}}^{t}{\nu} = {\mathit{X}}^{t}{\mathit{X}}{\varphi} = {\mathit{K}}{\varphi},
\end{align}
where ${\mathit{K}}$ is a $n \times n$ symmetric kernel (Gram) matrix. Here $\mathit{K}_{ij} = \kappa({x}_{i},{x}_{j}) = \langle {x}_{i}, {x}_{j} \rangle$ represents a simple inner product kernel, but can be replaced by \eqref{eq:kernelmix} by utilizing the kernel trick. 
Multiplying ${\mathit{X}}^{t}$ on both sides of \eqref{eq:keig}, results in the following generalized eigenvalue problem.
\begin{align}
{\mathit{K}}{\mathit{W}}^{(\text{b})}{\mathit{K}}{\varphi} &= \lambda{\mathit{K}}{\mathit{W}}^{(\text{w})}{\mathit{K}}{\varphi}.
\label{eq:kgen}
\end{align}
		
	Let ${\Psi}=\{{\varphi}_{k}\}_{k=1}^{r}$ be the $r$ generalized eigenvectors associated with the $r$ smallest eigenvalues $\lambda_{1} \le \lambda_{2}, \ldots, \le \lambda_{r}$. A test sample ${x}_{\textit{test}}$ can be embedded in $\mathcal{H}$ via
	\begin{align}
		({\mathit{X}}{\Psi})^{t}{x}_{\textit{test}} = {\Psi}^{t}{\mathit{X}}^{t}{x}_{\textit{test}} = {\Psi}^{t}{\mathit{K}}_{{\mathit{X}},{x}_{\textit{test}}},
		\label{eq:ckladaeig}
	\end{align}
	where ${\mathit{K}}_{{\mathit{X}},{x}_{\textit{test}}}$ is a $n \times 1$ vector. 	
Composite Kernel Local ADA (CKLADA) can likewise be implemented by replacing the weight matrices (${\mathit{W}}^{(\text{w})}$ and ${\mathit{W}}^{(\text{b})}$) above with their local counterparts defined in \eqref{eq:lada_w} and \eqref{eq:lada_b}. 

We note that in the proposed approach, the empirical kernel (Gram) matrix from \eqref{eq:kernelmix} that is formed as a weighted linear combination over all sources is used in the generalized eigenvalue problem for CKLADA \eqref{eq:ckladaeig}. The algorithm projects the data from $M$ sources onto a \emph{unified RKHS} through a bank of kernels individually optimized for each source. The final embedding seeks to optimally separate (in an angular sense) data in the RKHS. The linear mixture of kernel enables us to optimize each kernel (for example the kernel parameters) for each source instead of applying a single kernel for all sources, and to specify source importance (via mixing weights) to the overall analysis task at hand. 

\emph{Practical Considerations:} We note the following free parameters in the overall embedding that affect the subspace that is generated: Embedding dimension, $r$, mixture weights used in the composite kernel, $\{\alpha_m\}_{m=1}^M$, choice of kernel and related kernel parameters. We note that unlike some other embedding techniques such as LDA and its variants where the embedding dimension is upper bounded due to rank deficiency of the between class scatter, with composite kernel local ADA, the between class angular scatter is not rank limited, and as a result, the projection matrix resulting from the solution to the generalized eigenvalue problem does not enforce an upper bound on the embedding dimension. Hence, $r$ is a free parameter that represents the \emph{unified subspace} generated by all $M$ sources. The choice of $r$ should hence be governed by the information content (as quantified for example in the eigenspectra of the decomposition). The choice of weights can be made through cross validation or techniques such as kernel alignment --- in our experience, there is often a very wide plateau over a range of values of the weights, and hence we chose to use simple cross validation to learn weights from our training data. We utilized a standard radial basis function (RBF) kernel for each source ($K_m$), but the kernel parameter (width of the RBF kernel) is optimized for each source individually via cross validation.

\emph{Classification:} We note that following a CKLADA projection, a simple classifier can be utilized for down-stream analysis. This follows from the observation that applying kernel projections while simultaneously ensuring preservation of angular locality will result in subspaces where class-specific data are compactly clustered. We validate and measure the efficacy of subspaces resulting from CKLADA by utilizing the following classifiers: (1) A K Nearest neighbor (KNN) classifier, (2) A Gaussian maximum-likelihood (ML) classifier, and a (3) sparse representation based classifier (SRC) \cite{PAMI2009_SRC}. While the choice of KNN and ML are obvious for subspaces formed by Kernel projections as noted in \cite{YP2015}, we make a remark on choice of SRC as an additional classifier to measure efficacy of subspaces --- this choice is motivated not only by the observation that SRC has emerged as a powerful classification approach for high dimensional remote sensing data \cite{Cui2014,yuan2012visual,chen2011hyperspectral,Nasser2011SRC,zhang2011sparse} and that it exploits the inherent sparsity when representing samples using training dictionaries, but also because popular solvers used (e.g. Orthogonal Matching Pursuit, OMP) are driven by inner products to learn the sparse representation and hence they essentially exploit angular information. They implicitly seek a representation where a test sample is represented sparsely in a dictionary of training data such that the atoms that eventually contribute (have non-zero, significant representation coefficients) to the representation are angularly similar to the test data samples. We hence contend that CKLADA is particularly well suited for SRC and its variants.
	
\section{Experimental Setup and Results}
\label{sec:results}

\subsection{Dataset}

\begin{figure*}[ht] 
	\centering
	\includegraphics[width=15cm]{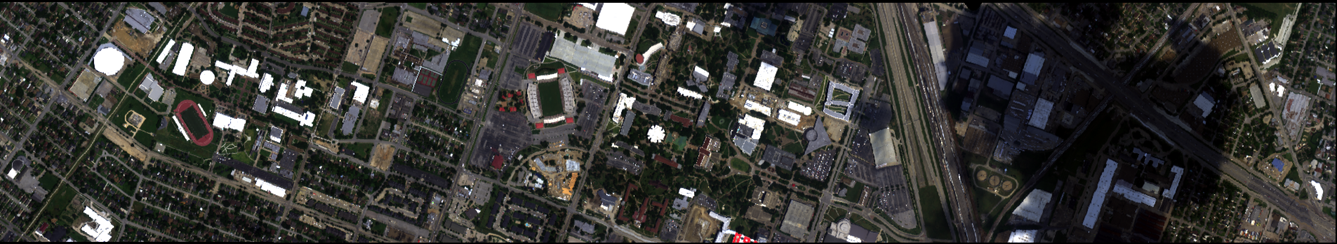} \\
	\includegraphics[width=17cm]{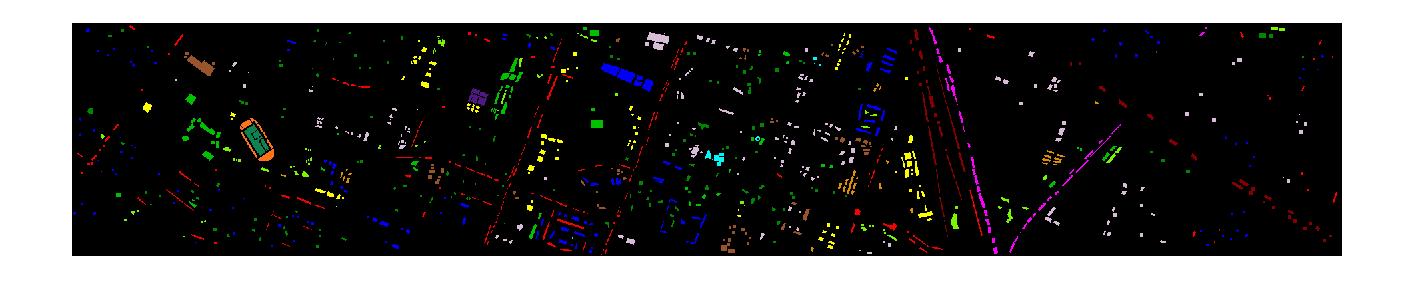} \\
	\includegraphics[width=15cm]{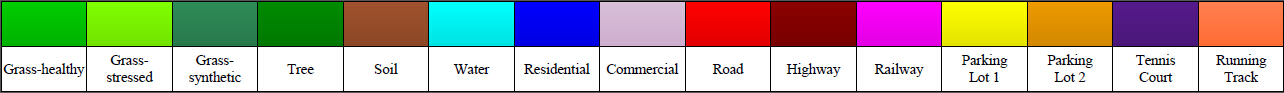}
	\caption{
		True color image of hyperspectral University of Houston data, and the ground truth.}
	\label{fig:img_uh}
\end{figure*}

The dataset we utilize represents a sensor fusion scenario, comprising of LiDAR pseudo-waveforms, and a hyperspectral image cube, and is popular in the remote sensing community as a benchmark. The data were acquired over the University of Houston campus and the neighboring urban area. All the images are at the same spatial resolution of 2.5 m and have the same spatial size of $349 \times 1340$. The hyperspectral image was acquired with the ITRES CASI sensor, containing 144 spectral bands, ranging from 380 nm to 1050 nm. The LiDAR DSM data was acquired by an Optech Gemini sensor and then co-registered to the hyperspectral image. The laser pulse wavelength and repetition rate were 1064 nm and 167 kHz, respectively. The instrument can make up to 4 range measurements. The total number of ground reference samples is 2832, covering 15 classes of interest, with approximately 200 samples for each class --- these were determined by photo-interpretation of high resolution optical imagery.. The groundtruth map is overlaid with the gray scale image showing one channel of the hyperspectral image in Fig. \ref{fig:img_uh}.
From the dense LiDAR point cloud, a pseudo-waveform was generated for each geolocation, that is co-registered with the hyperspectral image. The pseudo-waveform was generated by quantizing the elevation into uniform sized bins, and determining the average intensity of points as a function of elevation bins. This provides us with a co-registered cube of waveform-like LiDAR data that is coregistered with our hyperspectral image. We note that like spectral reflectance profiles that have unique shapes depending on the material in the pixel, shapes of pseudo-waveform also correlate with the material and topographic properties in the image. Hence, angular measures (such as provided by CKLADA) would be appropriate for such analysis compared to Euclidean measures.

\subsection{Baselines}
To validate the efficacy of the subspaces generated by CKLADA and CKADA, we setup classification experiments using the University of Houston multi-sensor dataset. We used popular and commonly employed embeddings as baselines to compare against, including CKLFDA and KPCA. Each of these embeddings was used with 3 classifiers: KNN, ML and SRC. We note that CKLFDA is a composite kernel counterpart of LFDA based on Euclidean distance measures, and is the best possible multi-source embedding that can be compared with CKLADA --- a comparison of CKLFDA vs. CKLADA provides a direct understanding of the benefits of angular information for multi-source embeddings, and of the resulting algorithmic framework proposed in sec. \ref{sec:proposed}. 

With CKLFDA, CKADA and CKLADA, we treat each sensor (hyperspectral imagery and pseudo-waveform LiDAR) as a source, each getting its dedicated base kernel. With a single kernel KPCA, we stack features from the two sensors and project them via a single transformation based on these methods. In all cases, we use RBF kernels as our base kernels, and the width of the kernel is determined via cross validation. Other free parameters including sparsity level used in SRC, number of nearest neighbors in $K-NN$ are also determined empirically from the training data via cross-validation.

\subsection{Visualization of Embeddings}

\begin{figure*}[ht]
  \centering
  \begin{subfigure}{\linewidth}
    \centering
	\includegraphics[height=4cm]{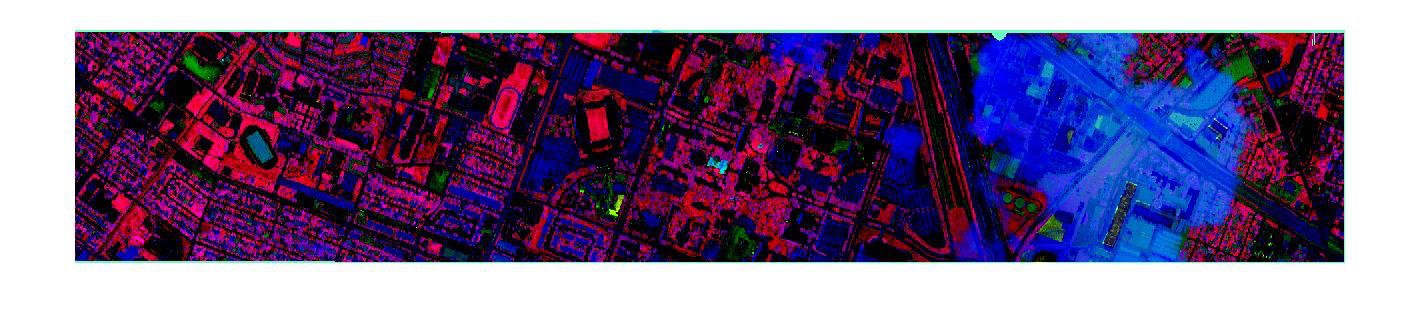} \\
    \caption{CKLADA (Proposed)}
  \end{subfigure}

  \begin{subfigure}{\linewidth}
    \centering
	\includegraphics[height=4cm]{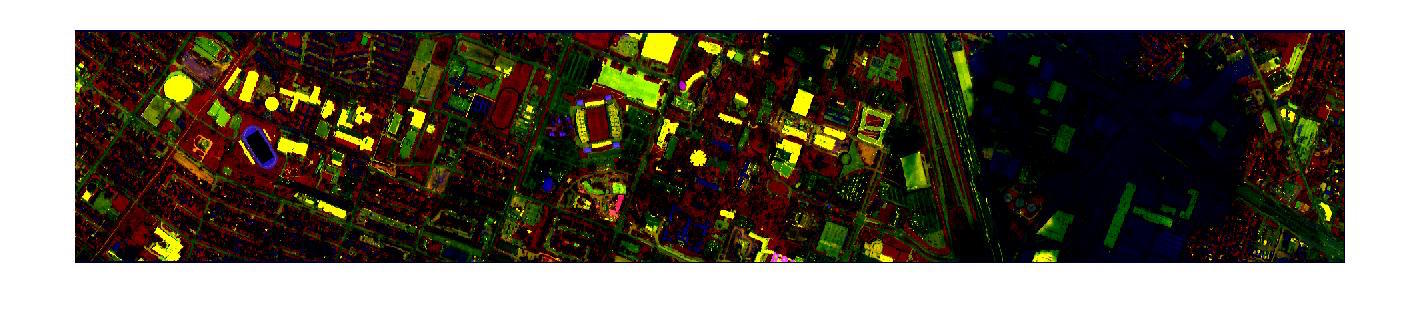} \\
    \caption{CKADA (Proposed)}
  \end{subfigure}  
  
  \begin{subfigure}{\linewidth}
    \centering
	\includegraphics[height=4cm]{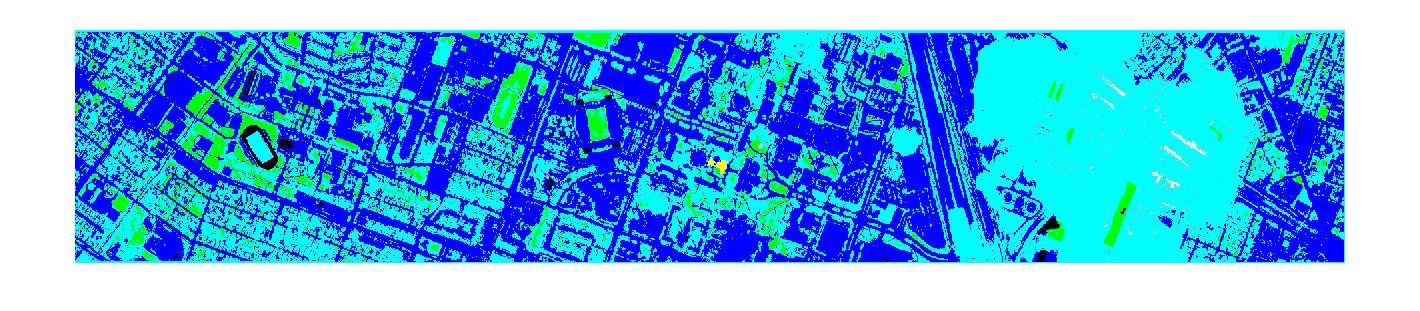}
    \caption{CKLFDA \cite{YP2015}}
  \end{subfigure}  
  \caption{Visualizing the projections as false color images resulting from projecting the multi-sensor data onto the first 3 significant eigenvectors using (a) CKLADA (proposed); (b) CKADA (proposed); and (c) CKLFDA.}  
\label{fig:img_vis}
\end{figure*}

To provide a visual demonstration on the power of composite kernel angular discriminant analysis for geospatial image analysis, we provide visualization of composite kernel projections CKLADA, CKADA (both angular discriminant analysis) and CKLFDA. These results are depicted in fig. \ref{fig:img_vis}. The figure depicts false color images generated by projecting the multi-sensor data onto the first three most significant eigenvectors learned from CKLADA, CKADA and CKLFDA respectively. It can be clearly seen that CKLADA (and CKADA to some degree) preserve object specific properties throughout the image (for example, the highly textured objects such as urban vegetation, residential areas etc. have their spatial context significantly preserved in the lower dimensional subspace). On the contrary, CKLFDA, which can be considered as the closest benchmark/baseline competitor does not perform as well. Further, towards the right corner of the image, we point the reader to the substantial benefit of CKLADA under cloud shadows - spatial structures under cloud shadows are visible under CKLADA (and CKADA to some extent), but not when using CKLFDA.

\subsection{Comparative Results}
\begin{figure*}[ht]
  \centering
  \begin{subfigure}{\linewidth}
    \centering
	\includegraphics[width=16.4cm]{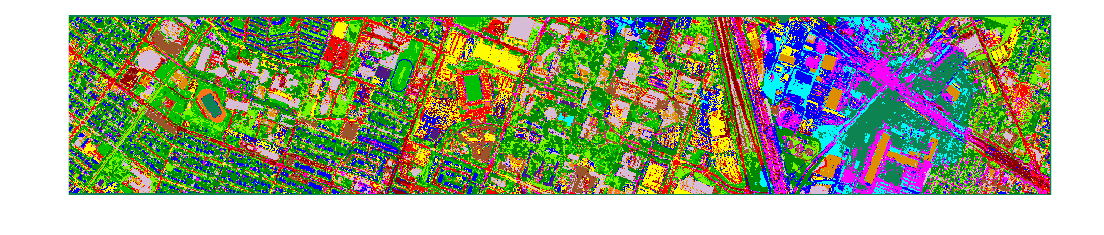} \\
    \caption{Classification map generated using CKLADA (Proposed) with 20 training samples per class.}
  \end{subfigure}

  \begin{subfigure}{\linewidth}
    \centering
	\includegraphics[width=16.2cm]{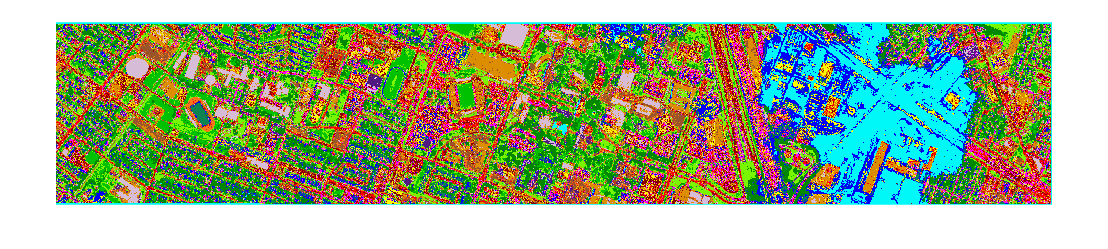} \\
	\caption{Classification map generated using CKLFDA \cite{YP2015} with 20 training samples per class.} 
	\vspace{12pt}
	\includegraphics[width=15cm]{color_bar.jpg}
  \end{subfigure}  
  
  \caption{Classification maps obtained from the multi-source (hyperspectral and pseudo-waveform LiDAR) dataset using (a) the proposed CKLADA embedding and (b) a CKLFDA (bottom) embedding, both with an SRC classifier following the embedding, and with 20 samples per class for training.}  
\label{fig:map_uh}
\end{figure*}

Experimental results comparing performance of CKADA and CKLADA with baseline embeddings are provided. As mentioned previously, free parameters were determined empirically via cross-validation. 
 Tab. \ref{tab:res_ss} depicts overall accuracy as a function of training sample size, ranging from a small to a sufficiently large value for the proposed and baseline embeddings with various classifiers. We notice that the proposed composite kernel angular discriminant analysis approaches (CKADA and CKLADA) provide anywhere from $5-10\%$ improvement in performance compared to state of the art (CKLFDA), and provide even higher accuracies compared to a traditional single-kernel baseline, KPCA.  We note that even when using very limited training data, CKLADA is able to substantially outperform other composite kernel and single-kernel methods (using just 10 samples per class, we obtain as much as a $10\%$ gain in performance with CKLADA). Even when using a sufficiently large training sample set (e.g. 50 samples per class), CKLADA and CKADA outperform other methods. 
 In Tab. {\ref{tab:res_src}, Tab. \ref{tab:res_ml}, and Tab. \ref{tab:res_knn},} we depict class specific accuracies, overall and average accuracies using the proposed methods and baselines using SRC, ML and KNN classifiers respectively. Once again, it is clear that CKADA and CKLADA consistently provide robust classification, particularly for the ``difficult'' classes (such as residential buildings, commercial buildings, roads, parking lots etc.). We also provide classification maps in fig. \ref{fig:map_uh} using the proposed method (CKLADA) and its closest competitor, CKLFDA, using the SRC classifier. We note that CKLADA results in a map with very little noise and misclassifications, and is particularly robust under the very challenging area in the right corner of the image that is under a cloud shadow (for e.g., when using CKLFDA, the area under a cloud shadow get systematically misclassified as water --- something that is visibly remedied by CKLADA). The improvements to misclassifications occurring over difficult classes is even more apparent with these ground cover classification maps.

\begin{table*}[ht]
\centering
\caption{Comparison of various feature embedding algorithms for multi-sensor image analysis as a function of training sample size}
\begin{tabular}{c c c c c c}
\hline
\multirow{2}{*}{Method} & \multicolumn{5}{c}{Number of training samples per class} \\
& 10 & 20 & 30 & 40 & 50\\
\hline
CKLADA-SRC & 73.2$\pm$3.8 & 87.1$\pm$1.1 & 90.4$\pm$1.1 & 92.3$\pm$0.9 & 93.3$\pm$0.9 \\
CKADA-SRC & 74.5$\pm$2.3 & 85.8$\pm$1.2 & 89.2$\pm$1.2 & 91.2$\pm$1.1 & 92.3$\pm$0.8 \\
CKLFDA-SRC & 66.6$\pm$2.5 & 81.6$\pm$1.6 & 86.2$\pm$1.3 & 87.8$\pm$1.0 & 88.9$\pm$1.0 \\
KPCA-SRC & 67.9$\pm$1.4 & 79.3$\pm$1.3 & 84.1$\pm$1.1 & 86.7$\pm$0.8 & 88.5$\pm$0.9 \\
CKLADA-ML & 74.33$\pm$2.4 & 85.7$\pm$1.6 & 91.1$\pm$1.2 & 93.3$\pm$1.0 & 94.3$\pm$0.7 \\
CKADA-ML & 77.2$\pm$2.7 & 86.7$\pm$1.6 & 91.6$\pm$1.2 & 93.0$\pm$1.0 & 94.0$\pm$0.8 \\
CKLFDA-ML & 70.4$\pm$2.4 & 81.2$\pm$1.7 & 86.7$\pm$1.6 & 88.9$\pm$1.1 & 90.2$\pm$1.0 \\
KPCA-ML & 72.15$\pm$2.7 & 85.1$\pm$1.9 & 91.2$\pm$1.2 & 93.3$\pm$1.0 & 94.3$\pm$0.8 \\
CKLADA-KNN & 80.3$\pm$1.7 & 88.1$\pm$1.3 & 91.4$\pm$1.0 & 93.0$\pm$0.9 & 93.9$\pm$0.7 \\
CKADA-KNN & 80.3$\pm$1.6 & 87.0$\pm$1.2 & 90.1$\pm$1.0 & 91.4$\pm$0.9 & 92.4$\pm$0.8 \\
CKLFDA-KNN & 70.7$\pm$2.2 & 82.5$\pm$1.6 & 86.5$\pm$1.2 & 88.1$\pm$1.0 & 89.3$\pm$0.9 \\
KPCA-KNN & 69.7$\pm$1.6 & 79.2$\pm$1.1 & 83.7$\pm$1.3 & 86.4$\pm$0.9 & 87.8$\pm$0.9 \\
\hline
\end{tabular}
\label{tab:res_ss}
\end{table*}

\begin{table*}[ht]
\centering
\caption{Using proposed and baseline feature embeddings with SRC}
\begin{tabular}{c||c|c||c c c c}
\hline
\hline

\multirow{2}{*}{Class} & \multicolumn{2}{c||}{Samples} & \multicolumn{4}{c}{Methods} \\
&Train & Test & CKLADA & CKADA & CKLFDA & KPCA \\
\hline
\hline

Grass-healthy & 30 & 168 & 99.4$\pm$1.0 & 99.0$\pm$2.0 & 98.6$\pm$1.7 & 98.0$\pm$1.8 \\ \cline{2-3}
Grass-stressed & 30 & 160 & 97.7$\pm$1.4 & 96.0$\pm$1.6 & 98.0$\pm$1.2 & 95.9$\pm$3.1 \\ \cline{2-3}
Grass-synthetic & 30 & 162 & 99.7$\pm$0.5 & 99.7$\pm$0.4 & 95.8$\pm$2.5 & 98.3$\pm$1.3 \\ \cline{2-3}
Tree & 30 & 158 & 98.1$\pm$1.2 & 95.9$\pm$2.7 & 98.8$\pm$1.2 & 99.5$\pm$0.5 \\ \cline{2-3}
Soil & 30 & 156 & 99.8$\pm$0.3 & 98.6$\pm$0.9 & 98.5$\pm$1.7 & 97.3$\pm$2.4 \\ \cline{2-3}
Water & 30 & 152 & 98.7$\pm$2.3 & 95.1$\pm$3.0 & 97.0$\pm$2.3 & 96.9$\pm$2.3 \\ \cline{2-3}
Residential & 30 & 166 & 85.8$\pm$5.5 & 81.3$\pm$5.0 & 80.3$\pm$4.6 & 77.1$\pm$6.1 \\ \cline{2-3}
Commercial & 30 & 161 & 86.8$\pm$5.5 & 82.9$\pm$6.4 & 77.2$\pm$6.4 & 79.5$\pm$5.6 \\ \cline{2-3}
Road & 30 & 163 & 80.2$\pm$5.7 & 78.5$\pm$6.1 & 73.7$\pm$5.5 & 64.4$\pm$5.1 \\ \cline{2-3}
Highway & 30 & 161 & 90.9$\pm$3.2 & 90.8$\pm$3.7 & 86.9$\pm$3.9 & 77.9$\pm$5.2 \\ \cline{2-3}
Railway & 30 & 151 & 88.5$\pm$3.9 & 86.2$\pm$4.5 & 82.6$\pm$4.1 & 76.1$\pm$5.8 \\ \cline{2-3}
Parking Lot 1 & 30 & 162 & 75.2$\pm$5.7 & 74.2$\pm$6.3 & 67.8$\pm$6.4 & 62.3$\pm$5.0 \\ \cline{2-3}
Parking Lot 2 & 30 & 154 & 65.5$\pm$4.7 & 74.5$\pm$4.5 & 58.4$\pm$6.8 & 50.7$\pm$5.7 \\ \cline{2-3}
Tennis Court & 30 & 151 & 99.5$\pm$0.5 & 98.5$\pm$1.1 & 98.6$\pm$2.0 & 96.6$\pm$2.2 \\ \cline{2-3}
Running Track & 30 & 157 & 98.9$\pm$0.5 & 98.7$\pm$0.7 & 98.4$\pm$0.9 & 99.8$\pm$0.4 \\ \cline{2-3}
\hline
\hline
OA & -- & -- & 91.0$\pm$1.0 & 89.9$\pm$0.9 & 87.3$\pm$1.2 & 84.7$\pm$1.1 \\
AA & -- & -- & 91.0$\pm$2.8 & 90.0$\pm$3.2 & 87.4$\pm$3.4 & 84.7$\pm$3.5 \\
\hline
\end{tabular}
\label{tab:res_src}
\end{table*}

\begin{table*}[ht]
\centering
\caption{Using proposed and baseline feature embeddings with Gaussian ML}
\begin{tabular}{c||c|c||c c c c}
\hline
\hline

\multirow{2}{*}{Class} & \multicolumn{2}{c||}{Samples} & \multicolumn{4}{c}{Methods} \\
&Train & Test & CKLADA & CKADA & CKLFDA & KPCA  \\
\hline
\hline
Grass-healthy & 30 & 168 & 94.9$\pm$5.4 & 95.1$\pm$4.1 & 92.6$\pm$5.2 & 95.9$\pm$4.4 \\ \cline{2-3}
Grass-stressed & 30 & 160 & 99.4$\pm$0.8 & 97.8$\pm$1.9 & 99.8$\pm$0.3 & 98.6$\pm$1.9 \\ \cline{2-3}
Grass-synthetic & 30 & 162 & 96.5$\pm$0.3 & 96.8$\pm$2.7 & 88.2$\pm$4.8 & 96.7$\pm$2.7 \\ \cline{2-3}
Tree & 30 & 158 & 99.7$\pm$0.8 & 99.5$\pm$0.8 & 98.9$\pm$1.7 & 99.9$\pm$0.3  \\ \cline{2-3}
Soil & 30 & 156 &  97.6$\pm$2.7 & 93.2$\pm$4.4 & 95.0$\pm$4.4 & 97.4$\pm$2.5 \\ \cline{2-3}
Water & 30 & 152 &  95.1$\pm$2.9 & 92.6$\pm$4.0 & 94.0$\pm$3.5 & 96.0$\pm$2.8  \\ \cline{2-3}
Residential & 30 & 166 & 86.1$\pm$6.9 & 80.8$\pm$6.9 & 82.4$\pm$7.9 & 82.2$\pm$7.8  \\ \cline{2-3}
Commercial & 30 & 161 & 86.1$\pm$8.3 & 91.3$\pm$6.6 & 73.5$\pm$11.8 & 86.5$\pm$9.7  \\ \cline{2-3}
Road & 30 & 163 & 83.5$\pm$8.6 & 87.8$\pm$6.8 & 76.8$\pm$7.2 & 82.6$\pm$8.9 \\ \cline{2-3}
Highway & 30 & 161 & 82.8$\pm$8.3 & 83.9$\pm$6.9 & 76.7$\pm$7.8 & 83.4$\pm$8.3  \\ \cline{2-3}
Railway & 30 & 151 & 84.5$\pm$7.7 & 81.8$\pm$7.6 & 80.6$\pm$8.0 & 83.6$\pm$8.6  \\ \cline{2-3}
Parking Lot 1 & 30 & 162 & 71.7$\pm$8.7 & 49.8$\pm$11.6 & 64.5$\pm$7.6 & 71.8$\pm$10.2 \\ \cline{2-3}
Parking Lot 2 & 30 & 154 & 92.2$\pm$3.3 & 96.7$\pm$1.5 & 84.8$\pm$6.7 & 92.5$\pm$4.3  \\ \cline{2-3}
Tennis Court & 30 & 151 & 98.0$\pm$2.0 & 96.7$\pm$3.4 & 95.5$\pm$2.8 & 98.5$\pm$1.9  \\ \cline{2-3}
Running Track & 30 & 157 & 96.6$\pm$2.0 & 97.1$\pm$1.9 & 94.9$\pm$2.9 & 97.9$\pm$1.5 \\ \cline{2-3}
\hline
\hline
OA & -- & -- & 90.9$\pm$1.2 & 89.3$\pm$1.3 & 86.5$\pm$1.3 & 84.6$\pm$1.6 \\
AA & -- & -- & 91.0$\pm$4.7 & 89.4$\pm$4.7 & 86.6$\pm$5.5 & 90.9$\pm$5.0  \\
\hline
\end{tabular}
\label{tab:res_ml}
\end{table*}

\begin{table*}[ht]
\centering
\caption{Using proposed and baseline feature embeddings with KNN}
\begin{tabular}{c||c|c||c c c c}
\hline
\hline

\multirow{2}{*}{Class} & \multicolumn{2}{c||}{Samples} & \multicolumn{4}{c}{Methods} \\
&Train & Test & CKLADA & CKADA & CKLFDA & KPCA \\
\hline
\hline
Grass-healthy & 30 & 168 & 98.6$\pm$3.0 & 99.4$\pm$2.3 & 97.4$\pm$2.9 & 98.2$\pm$2.8 \\ \cline{2-3}
Grass-stressed & 30 & 160 & 97.5$\pm$1.5 & 96.9$\pm$1.4 & 97.3$\pm$2.0  & 97.2$\pm$2.6  \\ \cline{2-3}
Grass-synthetic & 30 & 162 & 99.5$\pm$0.8 & 99.7$\pm$0.4 & 96.9$\pm$1.9 & 97.9$\pm$1.6 \\ \cline{2-3} 
Tree & 30 & 158 & 98.1$\pm$1.8 & 96.0$\pm$2.7 & 96.9$\pm$2.9 & 99.7$\pm$0.4 \\ \cline{2-3}
Soil & 30 & 156 & 99.7$\pm$0.5 & 98.5$\pm$0.8 & 98.1$\pm$2.6 & 98.1$\pm$1.1 \\ \cline{2-3}
Water & 30 & 152 & 98.4$\pm$2.4 & 94.8$\pm$2.4 & 97.3$\pm$2.2 & 96.4$\pm$1.7 \\ \cline{2-3}
Residential & 30 & 166 & 84.6$\pm$5.2 & 79.6$\pm$6.3 & 73.8$\pm$6.0 & 64.4$\pm$6.7  \\ \cline{2-3}
Commercial & 30 & 161 & 82.1$\pm$8.1 & 76.4$\pm$8.2 & 77.5$\pm$7.8 & 79.2$\pm$6.7 \\ \cline{2-3}
Road & 30 & 163 & 83.1$\pm$4.9 & 78.7$\pm$5.1 & 74.0$\pm$5.8  & 64.3$\pm$5.3  \\ \cline{2-3}
Highway & 30 & 161 & 94.3$\pm$2.7 & 93.3$\pm$3.2 & 89.0$\pm$3.4 & 80.8$\pm$3.8 \\  \cline{2-3}
Railway & 30 & 151 & 93.4$\pm$3.2 & 90.3$\pm$4.2 & 83.9$\pm$5.3 & 76.0$\pm$6.3 \\ \cline{2-3}
Parking Lot 1 & 30 & 162 & 78.8$\pm$5.9 & 76.6$\pm$6.4 & 70.7$\pm$6.5 & 67.8$\pm$6.5 \\ \cline{2-3}
Parking Lot 2 & 30 & 154 & 65.4$\pm$5.0 & 69.6$\pm$4.1 & 54.2$\pm$4.9  & 41.0$\pm$4.7  \\ \cline{2-3}
Tennis Court & 30 & 151 & 99.6$\pm$0.5 & 99.2$\pm$0.8 & 98.3$\pm$1.7 & 99.3$\pm$0.7  \\ \cline{2-3}
Running Track & 30 & 157 & 99.1$\pm$0.4 & 98.8$\pm$0.6 & 97.7$\pm$1.0 & 99.2$\pm$0.7 \\ \cline{2-3}
\hline
\hline
OA & -- & -- & 91.5$\pm$1.1 & 89.8$\pm$1.0 & 86.8$\pm$1.1 & 83.9$\pm$1.1 \\
AA & -- & -- & 91.5$\pm$3.1 & 89.8$\pm$3.3 & 86.7$\pm$3.8 & 84.0$\pm$3.4 \\
\hline
\end{tabular}
\label{tab:res_knn}
\end{table*}

\section{Conclusions}
\label{sec:Conclusions}
We presented a composite kernel variant of angular discriminant analysis and local angular discriminant analysis. Angular discriminant analysis was previously shown to be very beneficial for high dimensional hyperspectral classification. In this paper, we expanded those developments via a composite kernel and demonstrated that this paradigm can be a very useful feature embedding algorithm in multi-source scenarios, such as when fusing multiple geospatial images. We validated our results with a popular multi-sensor benchmark and demonstrated that composite kernel angular discriminant analysis consistently outperforms other feature embeddings. 

\bibliographystyle{IEEEtran}
\bibliography{string-defs,bibfile,reference_CKLFDA,/Users/sprasad/Dropbox/Research/BibFileSaurabh}

\end{document}